\pdfoutput=1

\documentclass[11pt]{article}

\usepackage[]{EMNLP2023}

\usepackage{times}
\usepackage{latexsym}

\usepackage[T1]{fontenc}

\usepackage[utf8]{inputenc}

\usepackage{microtype}

\usepackage{inconsolata}

\usepackage{algorithm}
\usepackage{algorithmicx}
\usepackage{algpseudocode}

\usepackage{multirow}
\usepackage{subcaption}
\usepackage{makecell}
\usepackage{array} 
\usepackage{mathtools}
\usepackage{amsmath}
\usepackage{url}
\usepackage{enumitem}
\usepackage{tabularx}
\usepackage{pdflscape}
\usepackage{lipsum}
\usepackage{booktabs}
\usepackage[toc,page]{appendix}
\usepackage{amsfonts}
\usepackage{bbm}
\newcolumntype{P}[1]{>{\centering\arraybackslash}p{#1}}
\newcolumntype{Q}[1]{>{\raggedright\arraybackslash}m{#1}}
\newcommand\redit[1]{\textcolor{red}{\textit{#1}}}
\usepackage{ulem}

\title{Style-News: Incorporating Stylized News Generation and Adversarial Verification for Neural Fake News Detection}

\author{Wei-Yao Wang, Yu-Chieh Chang, Wen-Chih Peng \\
  Department of Computer Science, National Yang Ming Chiao Tung University, Taiwan \\
  \texttt{sf1638.cs05@nctu.edu.tw, jessie86915@gmail.com, wcpengcs@nycu.edu.tw} \\
}
\begin{document}
\maketitle

\begin{abstract}
With the improvements in generative models, the issues of producing hallucinations in various domains (e.g., law, writing) have been brought to people’s attention due to concerns about misinformation.
In this paper, we focus on neural fake news, which refers to content generated by neural networks aiming to mimic the style of real news to deceive people.
To prevent harmful disinformation spreading fallaciously from malicious social media (e.g., content farms), we propose a novel verification framework, Style-News, using publisher metadata to imply a publisher's template with the corresponding text types, political stance, and credibility.
Based on threat modeling aspects, a style-aware neural news generator is introduced as an adversary for generating news content conditioning for a specific publisher, and style and source discriminators are trained to defend against this attack by identifying which publisher the style corresponds with, and discriminating whether the source of the given news is human-written or machine-generated.
To evaluate the quality of the generated content, we integrate various dimensional metrics (language fluency, content preservation, and style adherence) and demonstrate that Style-News significantly outperforms the previous approaches by a margin of 0.35 for fluency, 15.24 for content, and 0.38 for style at most.
Moreover, our discriminative model outperforms state-of-the-art baselines in terms of publisher prediction (up to 4.64\%) and neural fake news detection (+6.94\% $\sim$ 31.72\%).
\end{abstract}

\section{Introduction}
In recent years, social media have been used as platforms for people to share information due to non-distance on the Internet.
However, the amount of deceptive news has also increased from vicious social media such as content farms by changing some words from their templates; this problem has been widely tackled by detecting the veracity of the news \cite{DBLP:journals/csur/ZhouZ20}.
With the advancement of generative pre-trained models (e.g., \citet{DBLP:journals/corr/abs-2303-08774}), the issues of hallucinatory contents have been raised in various domains, e.g., scientific writing \cite{alkaissi2023artificial}, law \cite{forbes2023}.
In this paper, we focus on neural fake news, which has become an emerging societal crisis \cite{DBLP:conf/aaai/ShuLDL21,DBLP:conf/acl/FungTRPJCMBS20,DBLP:journals/corr/abs-2304-01487,reuters2023}, aiming to produce human-like news via AI models at scale to defraud humans \cite{DBLP:conf/acl/FungTRPJCMBS20}.
Therefore, it is crucial to develop verification techniques for defending against neural fake news\footnote{We follow \cite{DBLP:conf/nips/ZellersHRBFRC19} in using the term neural fake news to address machine-generated fake news.}.

The recent progress of neural fake news lies primarily in synthetic news generation. 
Early research on synthetic news generation relied on hand-written rules \cite{van2014journalist} or templates \cite{DBLP:conf/inlg/LeppanenMGT17}. 
With the proposed controllable text generation (CTG), text generation can be applied based on given attributes \cite{DBLP:journals/corr/abs-1909-05858, DBLP:journals/corr/abs-2201-05337}.
Grover \cite{DBLP:conf/nips/ZellersHRBFRC19} produces CTG on multi-field documents to create synthetic news, including domain, date, authors, headline, and body.
However, Grover neglects inherent factual discrepancies, which are tackled by retrieving external facts to enhance output consistency \cite{DBLP:conf/aaai/ShuLDL21}. 

\begin{figure}
  \centering
  \includegraphics[width=\linewidth]{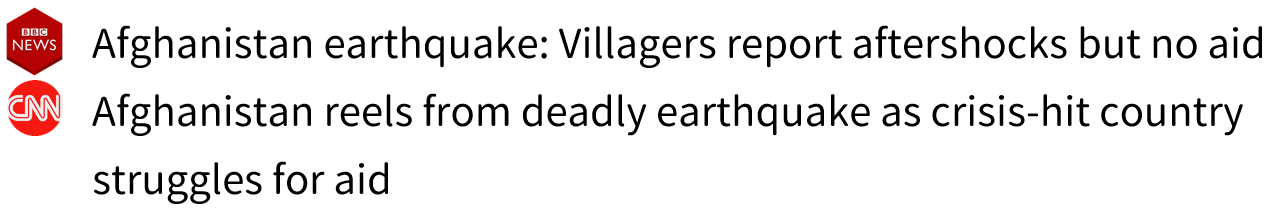}
  \caption{An example of news from different publishers.}
  \label{fig:example}
\end{figure}

Despite the above progress, there are two limitations in the previous work.
First, existing approaches to neural fake news detection fail to contemplate style information\footnote{We note that authors in \cite{DBLP:conf/nips/ZellersHRBFRC19} can be viewed as style information but are too sparse to learn the patterns.}.
In this paper, we focus on an unexplored facet of the style of news in neural news generation: \textbf{publisher} (e.g., CNN or BBC), which can be adopted as a template for vicious social media (e.g., content farms) to produce fake news that can attract readers to read news from the corresponding publisher \cite{baptista2020understanding}.
For example, news content, political stance, and social engagements will be influenced by hyper-partisan publishers.
Furthermore, different publishers are likely to describe an event with dissimilar content.
As shown in Figure \ref{fig:example}, we can observe that two publishers used different titles to describe the Afghanistan earthquake.
The former states the event with the format \textit{highlight: overview event}, whereas the latter uses a declarative sentence.
These can be viewed as templates from specific publishers, where malicious groups are able to produce fake news based on the templates to deceive readers who often read specific news.
Therefore, it becomes important to consider publisher information in synthetic neural news to detect it accurately before it is widely spread.

Second, previous work (e.g., \citet{DBLP:conf/nips/ZellersHRBFRC19,DBLP:conf/aaai/ShuLDL21}) evaluates the performance of defending neural fake news to classify the source of their generated news and the real news.
We argue that the discriminators are trained to distinguish generated text from the corresponding generators, which makes the evaluation process unfair due to the fitting discriminators.
It is essential to evaluate additional synthetic news that is not seen by models for fair comparisons.

In this work, we propose a novel framework, \textbf{Style-News}, with stylized news generation and two discriminators for publisher and neural fake news detection.
Stylized news generation (SNG) is introduced to utilize publisher information as an explicit style for controllably generating human-like news content.
To achieve fine-grained performance for SNG, the style discriminator is designed as a verifier for predicting the publisher of the generated content.
In addition, neural fake news detection (NFND) is proposed to enhance the accuracy of distinguishing human-written and machine-generated news, which can be viewed as a news verifier.
To tackle the second issue, we utilize the public dataset consisting of both synthetic and real news, VOA-KG2txt \cite{DBLP:conf/acl/FungTRPJCMBS20}, which is generated by a separate model, to fairly verify the capability and robustness of our neural fake news detection and other baselines. 
The contributions of this paper are summarized as follows:

\begin{itemize}
  \vspace{-5pt}
  \item 
  We propose an adversarial framework with a threat modeling perspective to address the publisher-faceted issue of neural fake news.
  Meanwhile, the stylized news generation incorporates publisher information to produce style-adherence and human-like news content.
  \item
  To compare neural fake news detection fairly, we propose a fair evaluation pipeline by using an additional dataset instead of self-generated data to evaluate the robustness of our model and baselines. To the best of our knowledge, our work is the first to conduct comprehensive experiments for neural news generation and detection, which benefits future researchers with multi-dimensional performance aspects.
  \vspace{-5pt}
  \item
  Extensive experiments show that our generator significantly outperforms on multiple general news datasets in terms of fluency, content, and style qualities.
  Moreover, Style-News achieves a new state-of-the-art result on the neural fake news tasks, which demonstrates the effectiveness of our defense framework.
\end{itemize}

\section{Related Work}
\label{related}
\noindent\textbf{Stylized text generation.}
Pre-trained language models (PLMs) have been widely adopted in various natural language tasks, which are trained on the large-scale corpus to have the ability to understand generic knowledge of text \cite{DBLP:journals/corr/abs-2105-10311}.
In recent years, generative PLMs have aimed to mimic the style of human beings to produce readable text from input prompts \cite{DBLP:journals/corr/abs-2105-10311}.
For instance, GPT-family \cite{radford2018improving,radford2019language,DBLP:conf/nips/2020,DBLP:journals/corr/abs-2303-08774} is a de facto generative model which achieves the robustness of text generation tasks.
Accordingly, we adopted GPT-2 as the generation backbone following previous work.

Most of the research on stylized text generation puts efforts into the psycholinguistic aspect such as formal and casual with supervised settings \cite{DBLP:conf/emnlp/WangWMLC19,DBLP:journals/corr/abs-1909-08349}.
However, supervised training requires a large amount of labeled data, which is difficult to generalize to practical tasks.
\citet{DBLP:conf/iclr/DathathriMLHFMY20} tackled this issue by integrating a PLM with attribute classifiers to construct controlled text generation without training on the language model.
StyleLM \cite{DBLP:conf/aaai/SyedVSNV20} pre-trains a Transformer-based masked language model and fine-tunes on an author-specific corpus using DAE loss.
However, the style of information has not been addressed in the neural news generation, which produces human-like news efficiently to deceive people based on existing publisher templates.

\noindent\textbf{Neural fake news detection.} 
The issues of fake news detection have been widely discussed since fake news covers a wide range of topics that may influence the public's views, political motives as well as social engagements \cite{DBLP:journals/sigkdd/ShuSWTL17}.
With the improvement of the generative PLM, \citet{DBLP:conf/nips/ZellersHRBFRC19} identified the problems of neural fake news and developed verification techniques by constructing controllable news generation as an adversary, and exploring potential defenses to mitigate the threats.
To tackle the limitations of contradiction or missing details between the generated news and input prompt, FactGen \cite{DBLP:conf/aaai/ShuLDL21} and InfoSurgeon \cite{DBLP:conf/acl/FungTRPJCMBS20} are introduced to improve the consistency of synthetic news by incorporating external knowledge. 
Nonetheless, previous work failed to explore style information to prevent neural fake news with specific templates, while we incorporate publisher information to generate style-aware news content to demonstrate the great potential of using style information and the awareness to defend against misinformation.

\section{Approach}
\label{method}
\begin{figure*}
  \centering
  \includegraphics[width=0.85\linewidth]{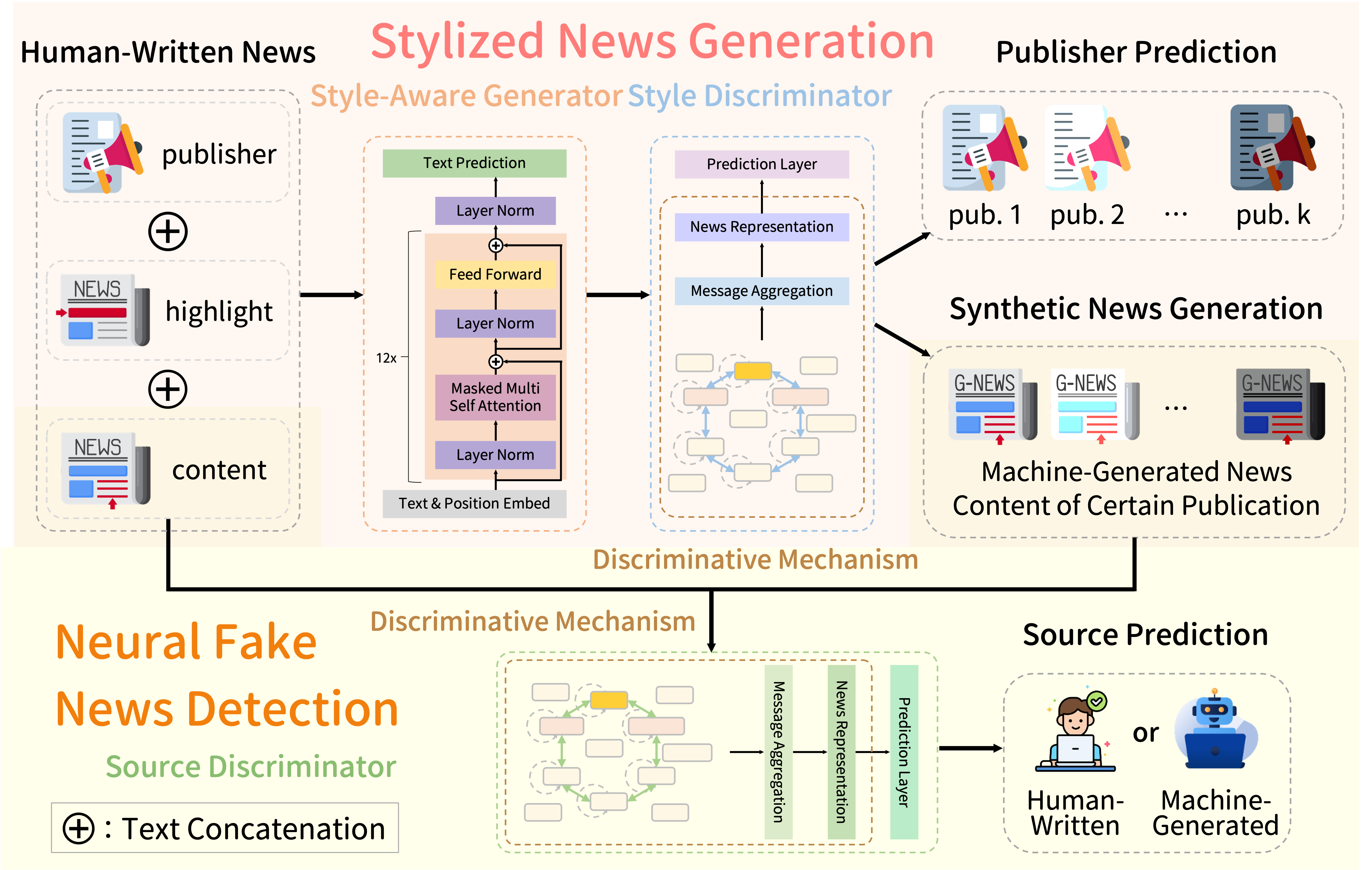}
  \caption{Illustration of the Style-News framework.}
  \label{fig:framework}
\end{figure*}

\subsection{Problem Statement}
\label{problem}
In this paper, we address the neural news detection problem in an adversarial setting similar to \cite{DBLP:conf/nips/ZellersHRBFRC19}.
We denote the \textbf{attack} phase as stylized neural news generation, and the \textbf{defense} phase as source discrimination.

In the attack phase, the input sequence contains news content, highlights, and publisher information.
The highlights can be a news title or summary based on the different datasets.
The goal of the generator $G_{style}$ is to produce news content that mimics the style of real news conditioning on a specified publisher, which cannot be distinguished by the source discriminator $D_{source}$.
The human-like news $N_M$ generated by $G_{style}$ can serve as potential threats that help the source discriminator learn to defend against neural fake news.

In the defense phase, the source discriminator $D_{source}$ aims to learn to distinguish if input news $N$ is human-written or machine-generated as:
\begin{equation}
    \label{eqn:3}
    D_{source}(N) \rightarrow y; y \in \{H, M\},
\end{equation}
where $H$ and $M$ denote human-written and machine-generated news, respectively.

\subsection{Style-News Framework}
\label{framework}
Figure \ref{fig:framework} presents the model architecture of Style-News, where a stylized news generation module aims to generate synthetic news with a style-aware generator and style discriminator by taking news title, summary, content, and publisher as inputs.
The neural fake news detection module classifies the source according to whether the input news content is human-written or machine-generated to enable the model to identify neural news.
With the adversarial training on the generator and source discriminator, we are able to build up a stronger generator to produce style-aware news content; meanwhile, we have designed a robust verification mechanism to detect neural fake news.
Detailed descriptions of the model are provided as follows.

\subsection{Stylized News Generation}
\label{generation}
The stylized news generation module aims to produce expressive news based on writing style, which has not been utilized in the previous work.
To generate stylized news, the style-aware generator is introduced by using publisher, title or summary, and content, and incorporates the style discriminator to reinforce the threat modeling aspect.

\noindent\textbf{Style-aware generator.}
\label{G_style}
Following \cite{DBLP:conf/nips/ZellersHRBFRC19,DBLP:conf/iclr/DathathriMLHFMY20}, we adopt GPT-2 \cite{radford2019language} as the generator backbone to produce news content.
However, GPT-2 cannot take news metadata (e.g., publisher) into account.
Therefore, we convert the token sequence of the publisher, highlight, and content as text prompts with task-context tokens as shown in Figure \ref{Input_Sequence}.
Formally, the prompt template of the input sentence is defined as:
\begin{flalign}
\label{eqn:1}
        &S = \textrm{\textless \textrm{|Start\_Publication|}\textgreater} \: publisher \: && \\\nonumber
        &\textrm{\textless |End\_Publication|\textgreater} \: highlight \: \textrm{\textless sep\textgreater} \: content, &&
\end{flalign}
where \textrm{\textless \textrm{|Start\_Publication|}\textgreater}, \textrm{\textless |End\_Publication|\textgreater} and \textrm{\textless sep\textgreater} are denoted as special tokens for indicating the publisher information, and separator tokens, respectively.
The special tokens $\textrm{\textless |Start\_Publication|\textgreater}$ and $\textrm{\textless |End\_Publication|\textgreater}$ enable the model to consider the importance of the publisher, which can also be controlled by different publishers during inferencing.
We truncate the input to $l$ tokens if the sequence length exceeds the maximum length.

During the training stage, we randomly separate human-written news from the training set $N_H$ into two groups for efficient training: the sampled group $N_H^{[sp]}$ and the unsampled group $N_H^{[usp]}$.
$N_H^{[usp]}$ is used to train the parameters of the generative model and $N_H^{[sp]}$ is used to generate the synthetic news for training the source discriminator.
Therefore, the input $N_H^{[usp]}$ contains news publisher, highlight, and content for the style-aware generator, and the objective function of $G_{style}$ is defined as a language model problem:
\begin{equation}
\label{eqn:gpt}
    \begin{aligned}
        L_{gen} = \sum_i(\log P(y_i|y_1,...,y_{i-1})).
    \end{aligned}
\end{equation}

\noindent\textbf{Discriminative mechanism (DM).}
\label{DM}
The goal of the DM is to capture the representation of given news and distinguish between corresponding classes to reinforce the model quality.
In the style discriminator $D_{style}$, the DM aims to identify which publisher the generated news belongs to.
In the source discriminator $D_{source}$, the DM attempts to classify the source into human-written or machine-generated (this will be discussed in $\S$\ref{discrimination}).

\begin{figure}
  \centering
  \includegraphics[width=\linewidth]{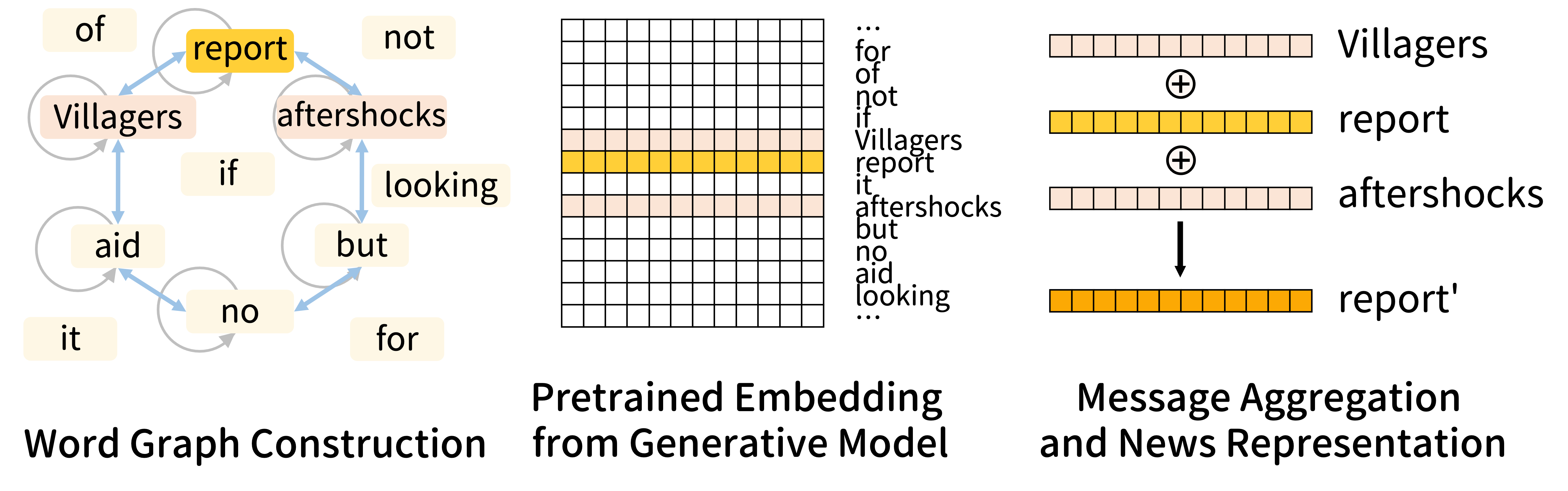}
  \caption{An example of the discriminative mechanism for \textit{Villagers report aftershocks but no aid}.}
  \label{word_graph}
\end{figure}

To capture the syntactic information, we propose a simple yet effective method by representing the input news (either human-written or machine-generated) in an inductive word graph as illustrated in Figure \ref{word_graph}; this approach has been utilized in various text classification tasks
\cite{DBLP:conf/acl/ZhangYCWWW20,DBLP:conf/coling/HuangCC22}.
Moreover, this design benefits our model generalizing to the unseen nodes compared with the common transductive graph models since the node embeddings in the word graph are initialized from the pre-trained word embeddings.
Specifically, each token is represented as a node in the word graph, and each token embedding is converted from the GPT-2 pretrained model.
Each node has two edges to connect with the former and latter tokens.
This graph construction procedure enables the model to not only recognize the common tokens of the input sequence but also to capture the contextual information between tokens.

Formally, the $i$-th node $w_{i}$ aggregate $p$ hops neighbor information to encode the contextual representations as $r_{w_i}'$:
\begin{equation}
    r_{w_i}'=(1-\alpha)\textrm{AGG}(\{r_{w_j}, w_j \in n(w_i)\})+\alpha r_{w_i},
\end{equation}
where $r_{w_j}$ is the node embedding, $n(w_i)$ is denoted as the $p$-hop neighborhood tokens of $w_i$, $\textrm{AGG}$ is the message aggregation with max pooling, and $\alpha \in \mathbb{R}^{1}$ is a trainable weight for adjusting the importance between the node itself and the neighbor.

After updating each node embedding, the news representation $r_N'$ is computed by aggregating node embeddings of news:
\begin{equation}
\label{eqn:rN}
r_N'=\sum_{w_i \in N}{r_{w_i}'},
\end{equation}
Finally, the news representation $r_N'$ is then fed into a linear layer to predict the label:
\begin{equation}
    \hat{y}=W' r_N' + b',
    \label{output_logits}
\end{equation}
where $W' \in \mathbb{R}^{d_r \times d_c}$ is a matrix that maps the news representation into the number of classes (i.e., publisher or news source) and $b' \in \mathbb{R}^{d_c}$ is the bias. 

To train the style discriminator, we minimize the cross-entropy loss $\mathcal{L}_{style}$:
\begin{equation}
    \mathcal{L}_{style} = - \sum_{i \in \{ pub 1, ..., pub k \} }{y_i \ln(\textrm{softmax}(\hat{y_i}))}.
    \label{minimize_style_loss}
\end{equation}


\subsection{Neural Fake News Detection}
\label{discrimination}
As the generator is capable of creating various types of news content based on the publisher, a source discriminator is introduced to prevent high-quality synthetic news from maliciously spreading and further misleading the public.

Specifically, the source discriminator $D_{source}$ adopts the same architecture as the DM.
The input is randomly sampled from either human-written $H$ or machine-generated $M$ (by Style-News) news content for the $D_{source}$.
We utilize the pretrained embeddings from $G_{style}$ to build the word graph and train $D_{source}$ by minimizing the cross-entropy loss for the class of news content:
\begin{equation}
    \mathcal{L}_{source} = - \sum_{i \in \{ H, M \} }{y_i \ln(\textrm{softmax}(\hat{y_i}))}.
    \label{minimize_source_loss}
\end{equation}

\subsection{Training Schedule}

To construct an adversarial structure, we establish Style-News in a nested loop and jointly train the style-aware generator with the style discriminator and the source discriminator respectively.
The training procedure is illustrated in Algorithm \ref{alg:train}, where we train $G_{style}$, $D_{style}$, and $D_{source}$ in order.
Generally, the inner loop is stylized news generation (Line~\ref{inner_start}-\ref{inner_end}), and the outer loop is neural fake news detection (Line~\ref{outer_start}-\ref{outer_end}), which is able to align the latent space of these modules and thus meet the goal of threat modeling.
We note that in the phase of training $D_{style}$ (Line~\ref{D_style_train_start}-\ref{D_style_train_end}), the input is human-written news in the first epoch to equip the ability for understanding news content of real publishers (Line~\ref{train_D_1}).
Afterwards, the input is the synthetic news generated by $G_{style}$ to detect the publisher of the generated news content (Line~\ref{train_D_2}).

\begin{algorithm}[t]
    \caption{Training procedure of Style-News.}
    \label{alg:train}
    \begin{algorithmic}[1]
    \Require The human-written news with publisher, title, and content $N_H$; first and second stage epoch number $epochs_{style}$ and $epochs_{source}$
    \Ensure Style-Aware Generator $G_{style}$, Style Discriminator $D_{style}$, and Source Discriminator $D_{source}$
    
    \Statex
    \item Initialize the $G_{style}$ from pretrained GPT-2 and add the special tokens; Initialize the parameters in $D_{style}$ and $D_{source}$ with Glorot uniform initializer \label{init}
    \item Randomly separate $N_H$ into the sampled and unsampled groups $N_H = \{N_H^{[sp]}, N_H^{[usp]}\}$.  \label{N_H_init}
    
    \For{$epoch_2$ = 1 to $epochs_{source}$} \label{outer_start}
        \State Train $G_{style}$ by $N_H^{[usp]}$ via maximizing $L_{gen}$ (Eq.\ref{eqn:gpt}) \label{inner_start}
            \For{$epoch_1$ = 1 to $epochs_{style}$} \label{D_style_train_start}
                \If{$epoch_2$ == 1} 
                    \State Train $D_{style}$ by $N_H^{[usp]}$ via minimizing $\mathcal{L}_{style}$ (Eq.\ref{minimize_style_loss}) \label{train_D_1}
                \Else 
                    \State Train $D_{style}$ by $N_M$ via minimizing $\mathcal{L}_{style}$ (Eq.\ref{minimize_style_loss}) \label{train_D_2}
                \EndIf
            \EndFor \label{D_style_train_end}
        \State Generate synthetic news $N_M$ \label{gen_N_S} from $N_H^{[sp]}$
        \item \Comment{Stylized News Generation} \label{inner_end}
    
        \State Concatenate $N_H$ and $N_M$ and randomly shuffle them as the input to the $D_{source}$ \label{D_source_data}
        \State Train $D_{source}$ via minimizing $\mathcal{L}_{source}$ (Eq.\ref{minimize_source_loss}) \label{D_source_train}
        \item \Comment{Neural Fake News Detection} \label{outer_end}
    \EndFor
    \end{algorithmic}
\end{algorithm}

\section{Experiments and Analysis}
\label{exp}

\begin{table*}[t]
    \small
    \centering 
    \scalebox{1.0}{

    \begin{tabular}{cc|cccccc}
    \toprule
    Criteria & Metric & CopyTransformer & GPT-2 & PPLM$_{\textrm{gen}}$ & Grover$_{\textrm{gen}}$ & FactGen & \textbf{Style-News} \\
    \midrule \midrule
    \multirow{2}{*}{Fluency} & Mauve $(\uparrow)$ & 0.7836 & 0.8050 & \underline{0.8827} & 0.8314 & 0.7836 & \textbf{0.8832} \\
     & Frontier $(\downarrow)$ & 0.9999 & 0.9300 & \underline{0.6634} & 0.7299 & 0.9999 & \textbf{0.6734} \\
    \midrule
    \multirow{2}{*}{Content} & SacreBLEU $(\uparrow)$ & 5.5527 & 8.1374 & \underline{14.7936}
 & 0.3084 & 13.1285 & \textbf{18.1064}  \\
     & MoverScore $(\uparrow)$ & 0.5166 & 0.5369 & 0.5217
 & 0.5010 & \underline{0.5434} & \textbf{0.5523} \\
    \midrule
    \multirow{2}{*}{Style} & Accuracy $(\uparrow)$ & 0.8918 & \underline{0.9273} & 0.5949 & 0.8378 & 0.7392 & \textbf{0.9609} \\
     & F1 $(\uparrow)$ & 0.5205 & 0.6898 & 0.5303 & \underline{0.8379} & 0.5000 & \textbf{0.8792} \\
     \midrule
     \multicolumn{2}{c|}{Avg. Rank} & 5.0 & \underline{3.3} & \underline{3.3} & 4.0 & 4.3 & \textbf{1.0} \\
    \bottomrule
    \end{tabular}
    }
\caption{Performance of synthetic news generation on CNN/DailyMail. The best result in each row is in boldface and the second best result is underlined.}
\label{tab:gen_cnndm}
\end{table*}

\subsection{Dataset}
We performed experiments on two news datasets that contain publisher metadata: CNN/DailyMail \cite{DBLP:conf/nips/HermannKGEKSB15,DBLP:conf/acl/SeeLM17} and All the News \cite{allthenews}.
The CNN/DailyMail dataset is written by journalists at CNN and the Daily Mail, and contains over 300,000 unique news articles and highlight sentences.
The training, validation, and testing sets are used as the official split.
The All the News dataset encompasses 143,000 articles and essays from 15 American publishers.
We picked the data with the common top 5 publishers (NPR, New York Post, Reuters, Washington Post, and Breitbart\footnote{Breitbart is known for publishing conspiracy theories, which can be further examined for the generation quality of the fake news publisher \cite{breitbart}.}) to ensure that the news of publishers has sufficient news to show the corresponding template patterns.

To defend against neural fake news, we follow \cite{DBLP:conf/nips/ZellersHRBFRC19,DBLP:conf/aaai/ShuLDL21} to test our source discriminator on machine-generated data.
However, previous evaluations only measured the effectiveness on their own self-generated datasets, which failed to measure the robustness of their discriminators.
Therefore, we utilized a public dataset containing both human-written and machine-generated news, VOA-KG2txt \cite{DBLP:conf/acl/FungTRPJCMBS20}, to fairly examine the discriminative performance of our models and the baselines.
VOA-KG2txt includes 15,000 real news articles from Voice of America and 15,000 neural fake news articles produced by the KG-to-text approach \cite{DBLP:conf/www/0004ZMK20}.
The testing set of these datasets is balanced.
The statistics of the datasets are described in Table \ref{tab:stat}.
All the results are the average of 5 random seeds.

\subsection{Implementation Details}
\label{implementation}
The word representation dimension $d_r$ is set to 768. 
For training the style-aware generator, we set the learning rate to $5\times$$10^{-5}$, warmup steps to 1000, and weight decay to 0.01.
The batch size in the training phase and generation phase was set to 2 and 32 respectively. 
For training the style discriminator and source discriminator, we used the Adam optimizer \cite{DBLP:journals/corr/KingmaB14} with an initial learning rate of $10^{-3}$, and weight decay was set to $10^{-4}$.
Dropout with a probability of 0.1 was applied after the linear layer. 
The max length of the token sequence $l$ was restricted to 1024.
The token would be converted to \textrm{\textless UNK\textgreater} special token if it does not match the dictionary of the pretrained model.
The number of hop $p$ is set to 1.
The $epochs_{style}$ and $epochs_{source}$ in Algorithm \ref{alg:train} were set to 10 and 5 respectively.
For the baseline models, we used default parameter settings as in their official implementations.
All the training and evaluation phases were conducted with Pytorch 1.7 on a machine with Ubuntu 20.04, Intel(R) Xeon(R) Silver 4110 CPU, and Nvidia GeForce RTX 2080 Ti GPU.

\begin{table*}[t]
    \small
    \centering
    \scalebox{1.0}{
    
    \begin{tabular}{cc|cccccc}
    \toprule
    Criteria & Metric & CopyTransformer & GPT-2 & PPLM$_{\textrm{gen}}$ & Grover$_{\textrm{gen}}$ & FactGen & \textbf{Style-News} \\
    \midrule \midrule
    \multirow{2}{*}{Fluency} & Mauve $(\uparrow)$ & 0.7849 & 0.8508 & 0.8707 &  \underline{0.8847} & 0.7756 & \textbf{0.8881} \\    & Frontier $(\downarrow)$ & 0.9966 & 0.7764 & 0.6865 & \underline{0.6642} & 1.0000 & \textbf{0.6467} \\
    \midrule
    \multirow{2}{*}{Content} & SacreBLEU $(\uparrow)$ & 0.3338 & 2.7831 & \underline{11.3883} & 1.0472 & 6.2606 & \textbf{14.5186} \\
    & MoverScore $(\uparrow)$ & 0.4942 & 0.5189 & \textbf{0.5519} & 0.5218 & 0.5304 & \underline{0.5448} \\
    \midrule
    \multirow{2}{*}{Style} & Accuracy $(\uparrow)$ & 0.2984 & 0.4721 & 0.5571 & \underline{0.5793} & 0.4828 & \textbf{0.5937} \\
    & F1 $(\uparrow)$ & 0.1478 & 0.3158 & \underline{0.4822} & 0.4733 & 0.4136 & \textbf{0.4921} \\
    \midrule
     \multicolumn{2}{c|}{Avg. Rank} & 5.7 & 4.3 & \underline{2.3} & 2.8 & 4.2 & \textbf{1.2} \\
    \bottomrule
    \end{tabular}

    }
\caption{Performance of synthetic news generation on All the News. The best result in each row is in boldface and the second best result is underlined.}
\label{tab:gen_atn}
\end{table*}

\subsection{Results of the Generative Models}
\noindent\textbf{Generative baselines.}
We selected 5 neural fake news generative baselines in this experiment to compare the generation quality of our Style-News.
Specifically, we compared CopyTransformer \cite{DBLP:conf/acl/SeeLM17}, GPT-2 \cite{radford2019language}, PPLM$_{\textrm{gen}}$ \cite{DBLP:conf/iclr/DathathriMLHFMY20}, Grover$_{\textrm{gen}}$ \cite{DBLP:conf/nips/ZellersHRBFRC19}, and FactGen \cite{DBLP:conf/aaai/ShuLDL21} for all the generative settings.

\noindent\textbf{Evaluation metrics.}
Since there is no existing work considering different facets of generation quality\footnote{We note that \citet{DBLP:conf/aaai/ShuLDL21} failed to consider the style aspect as evaluation, and the BLEU score is more suitable for content preservation instead of language fluency since repeated patterns have a larger score.},
we introduce three evaluation facets to assess the quality of generated news content: \textbf{language fluency:} Mauve score \cite{DBLP:conf/nips/PillutlaSZTWCH21} and Frontier Integral \cite{DBLP:conf/nips/LiuPWOCH21}, \textbf{content preservation:} SacreBLEU \cite{DBLP:conf/wmt/Post18} and MoverScore \cite{DBLP:conf/emnlp/ZhaoPLGME19}, and \textbf{style adherence:} RoBERTa-large with the training sets of CNN/DailyMail and All the News.
Detailed descriptions are introduced in Appendix \ref{appendix:generative-metrics}.

\noindent\textbf{Generation performance.}
Table \ref{tab:gen_cnndm} and Table \ref{tab:gen_atn} present the quality of the generation results in terms of language fluency (fluency), content presentation (content), and style adherence (style)\footnote{The generation samples are discussed in Appendix \ref{case}.}.
We can observe that Style-News consistently outperforms the generative baselines by a margin of 0.35 for fluency, 15.24 for content, and 0.38 for style at most for both datasets, which demonstrates the realistic-looking performance of our generated news content.
We summarize the observations as follows.

1) Using pre-trained models (i.e., GPT-2, PPLM$_{\textrm{gen}}$, Grover$_{\textrm{gen}}$, Style-News) to generate news content improves generation performance in terms of language fluency, which indicates the significance of incorporating prior knowledge from the pre-trained data.
2) We observed that the controllable text generative models (i.e., PPLM$_{\textrm{gen}}$, Grover$_{\textrm{gen}}$, Style-News) perform better on the style adherence aspect since non-controllable models fail to take the publisher information into account.
Therefore, our style-aware generator integrating publishers into the prompt to manipulate the style of news is superior to these baselines.
3) It is worth noting that all baselines are biased to generate human-like news content for only some facets, which indicates that they often focus on only specific aspects.
With the threat modeling design for the style-aware generator and style discriminator, our Style-News is capable of getting a human-like text with all criteria from the better-detected discriminator.

\subsection{Results of the Discriminative Models} 
\noindent\textbf{Discriminative baselines.}
To validate the performance of our proposed discriminators (including publisher and neural fake news detection), we further conducted experiments on publisher prediction and neural fake news detection with strong discriminative baselines: RoBERTa \cite{DBLP:journals/corr/abs-1907-11692}, PPLM$_{\textrm{def}}$ \cite{DBLP:conf/iclr/DathathriMLHFMY20}, Grover$_{\textrm{def}}$ \cite{DBLP:conf/nips/ZellersHRBFRC19}, GET \cite{DBLP:conf/www/XuWLWW22}, as well as CoCo \cite{DBLP:journals/corr/abs-2212-10341}. 
To investigate the effectiveness of feature-based methods, we follow the setting as \cite{DBLP:conf/coling/AichBP22} to add Linear Regression (LR), SVM, Ridge Regression (RR), KNN, and Random Forecast (RF) as machine learning baselines.

\noindent\textbf{Evaluation metrics.}
We adopt the common classification metric, macro F1 score, for measuring both publisher and neural fake news classifications.
We set the training epoch to 5 and selected the best model evaluating the validation set for all discriminative experiments.

\begin{figure}
  \centering
  \includegraphics[width=\linewidth]{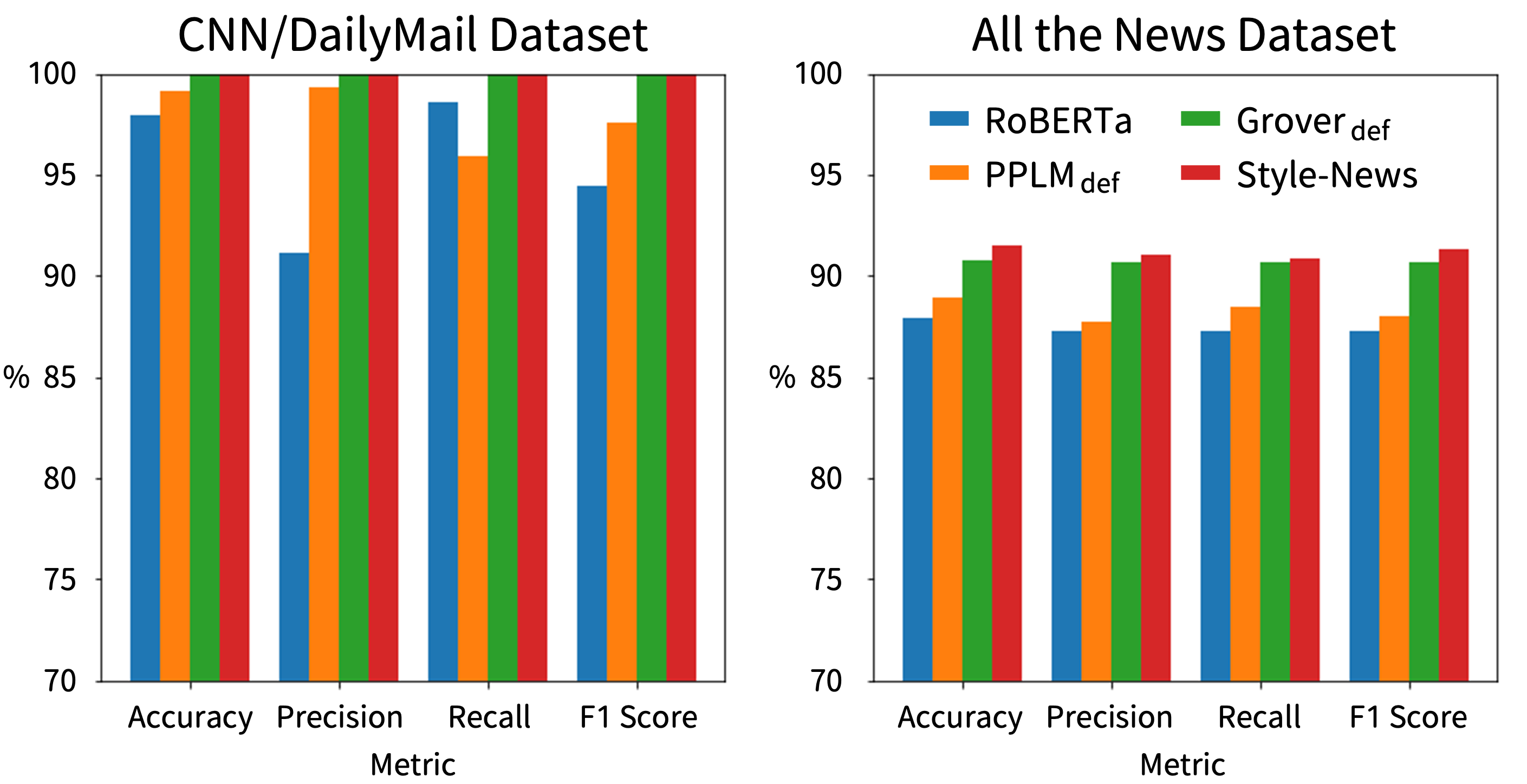}
  \caption{Performance of publisher prediction on CNN/DailyMail and All the News.}
  \label{publisher_prediction}
\end{figure}

\begin{table*}[]
    \small
    \centering
    \begin{tabular}{ccccc|ccccc|c}
    \toprule
    LR & SVM & RR & KNN & RF & RoBERTa & PPLM\textsubscript{def} & Grover\textsubscript{def} & GET & CoCo & Style-News \\
    \midrule
    72.35 & 75.84 & 68.07 & 81.10 & 80.67 & 92.15 & 89.83 & 74.82 & 82.45 & 93.82 & \textbf{98.55} \\
    \bottomrule
    \end{tabular}
    \caption{The F1 scores of neural fake news detection.}
    \label{VOA}
\end{table*}

\begin{table*}
    \small
    \centering
    \begin{tabular}{cc|cccc}
    \toprule
    Criteria & Metric & w/o Style w/o Source & w/o Style & w/o Source & \textbf{Style-News} \\
    \midrule \midrule
    \multirow{2}{*}{Fluency} & Mauve $(\uparrow)$ & 0.8045 & 0.8047 & 0.8164 & \textbf{0.8832} \\
    & Frontier $(\downarrow)$ & 0.9317 & 0.9310 & 0.8924 & \textbf{0.6734} \\
    \midrule
    \multirow{2}{*}{Content} & SacreBLEU $(\uparrow)$ & 8.1374 & 8.1405 & 7.9399 & \textbf{18.1064} \\
    & MoverScore $(\uparrow)$ & 0.5369 & 0.5371 & 0.5345 & \textbf{0.5523} \\
    \midrule
    \multirow{2}{*}{Style} & Accuracy $(\uparrow)$ & 0.9266 & 0.9253 & 0.9284 & \textbf{0.9609} \\
    & F1 Score $(\uparrow)$ & 0.6842 & 0.6822 & 0.7127 & \textbf{0.8792} \\
    \bottomrule
    \end{tabular}
\caption{Ablation study on CNN/DailyMail.}
\label{tab:ablation_CNNDM}
\end{table*}

\noindent\textbf{Publisher prediction.}
To examine the capability of distinguishing publishers of news, we carried out experiments with the discriminators to classify the publishers of real news.
Figure \ref{publisher_prediction} demonstrates the correctness of predicting the corresponding publisher given the news content\footnote{GET is neglected due to the gradient explosion while training on CNN/DailyMail and All the News.}.
Our Style-News outperforms the baselines by up to 4.64\% on All the News, and classifies perfectly on CNN/DailyMail, which demonstrates that jointly training the style-aware generator and style discriminator enables the model to understand the publisher template.
The observations are summarized as follows:

1) All models exhibit almost perfect performance on CNN/DailyMail since there are only two publishers, while the prediction gap becomes large on All the News.
2) RoBERTa hinders the performance compared with controllable generative models (i.e., PPLM$_{\textrm{def}}$, Grover$_{\textrm{def}}$, and Style-News), which indicates that additional information during training generators helps the corresponding discriminators to distinguish style information (e.g., publisher in this paper).
3) Both Grover$_{\textrm{def}}$ and Style-News achieve perfect performance on CNN/DailyMail but Style-News is superior to Grover$_{\textrm{def}}$ on All the News.
This comparison reveals the importance of considering publishers in the generator as well as using the discriminative mechanism.

\begin{table*}
    \small
    \centering
    \begin{tabular}{c|ccccc||c}
  \toprule 
    Criteria & CopyTransformer & GPT-2 & PPLM\textsubscript{gen} & Grover\textsubscript{gen} & FactGen & \textbf{Style-News} \\
    \midrule
    \midrule
    Language $(\uparrow)$ & 1.00 & 1.67 & 1.67 & \textbf{2.33} & 1.67 & \textbf{2.33} \\
    Content $(\uparrow)$ & 1.00 & 1.67 & 1.89 & 2.11 & 2.00 & \textbf{2.33} \\
    Style $(\uparrow)$ & 1.11 & 1.67 & 1.56 & 2.00 & 1.78 & \textbf{2.44} \\
    \bottomrule
    \end{tabular}

    \caption{The human evaluations of generated samples in both the CNN/DailyMail and the All the News datasets.}
    \label{tab:human-evaluation}
\end{table*}

\noindent\textbf{Neural fake news detection.}
To defend against synthetic fake news, we conducted experiments to examine the robustness of our Style-News.
We utilized VOA-KG2txt as the evaluated dataset to draw a fair comparison between our model and the baselines.
Table \ref{VOA} shows the performance of discriminative models on detecting the source of the input news content.
Specifically, our model surpasses all the baselines from 6.94\% to 31.72\%.
We conclude the observations as follows:

1) The models with the word graph (i.e., GET and Style-News) are superior to Grover$_{\textrm{def}}$, which verifies that the word graph can capture the syntactic meanings as structural information.
2) GET has substantially worse performance on neural fake news detection tasks since it takes claim-evidence interactions while there is no precise evidence of neural fake news in the real world.
In addition, Grover$_{\textrm{def}}$ suffers from degradation performance due to the sparsity learning from author information.
Attributed to capturing publisher style from the news, our model is thus able to effectively distinguish neural news.
3) All baselines degrade their performance in detecting neural news on the additional dataset, while our model consistently detects almost perfectly.
This suggests that evaluating the classification only on their generated news fails to measure the robustness of unseen news since the discriminators did not train on them.
Our model, in contrast, is still capable of classifying the news as either machine-generated or human-written, which can be used to not only defend against self-generated news but also against existing neural fake news.

\subsection{Result Analysis}
\noindent\textbf{Ablation study.}
To quantify the contributions of different discriminators of Style-News, we further conducted ablation experiments on CNN/DailyMail.
As shown in Table \ref{tab:ablation_CNNDM}, it is obvious that removing any discriminator (w/o Style and w/o Source) results in a significant performance drop in terms of all generated aspects.
Also, as expected, only using the generator leads to inferior performance in all metrics.
These results illustrate the reasonable and effective design of our model.
In addition, without the assistance of the style discriminator (w/o Style), the performance drops significantly in style adherence in terms of an F1 score of 0.21, indicating that the style discriminator can help enhance the ability to capture the writing style of the corresponding publisher.

\noindent\textbf{Human evaluation.}
We randomly sampled 3 generated news articles of each model from both CNN/DailyMail and All the News, which were annotated by 9 annotators without advanced knowledge of the source of the generated content to reflect real-world reader scenarios.
They were asked to evaluate the generated news in terms of language fluency, content preservation, and style adherence.
We provided some sample news from the corresponding publisher to let annotators evaluate style adherence.
The details of human evaluation questions were designed similarly to \cite{DBLP:conf/aaai/ShuLDL21}, i.e., the annotators should evaluate each question with a score of 1 (the worst) to 3 (the best).

Table \ref{tab:human-evaluation} lists the human evaluation results, which illustrate that our Style-News significantly outperforms all generative baselines in terms of all three aspects.
Quantitatively, our approach achieves 17\% and 37\% performance improvement over the best baseline in content preservation and style adherence, respectively.
This again reveals the enhancement of considering publisher information for stylized news generation.
 
\vspace{-3pt}
\section{Conclusion}
\label{conclusion}
\vspace{-4pt}
This paper presents Style-News, a novel adversarial framework to defend against the urgent neural fake news problem.
Distinct from existing generative models, our style-aware generator produces news with text prompts not only from news highlights and content, but also from publisher information, allowing the integration of additional metadata in the realm of text-metadata compositions.
Meanwhile, our neural fake news detection captures syntactic information by constructing the input as a graph for distinguishing the human-like news content.
Style-News sets new state-of-the-art results on both neural news generation and detection benchmarks with our comprehensive metrics.
We believe Style-News serves as a flexible framework for neural fake news detection, and multiple interesting directions could be further explored within the framework, such as prompt design, few-shot examples, etc.

\section{Ethics Considerations}
We discuss the potential usage and the potential risks of Style-News for ethical considerations.
\paragraph{Journalism assistants}
Inspired by \cite{DBLP:conf/aaai/ShuLDL21} who discussed helping journalists with claim generation using their fact retrieval mechanism, our method can provide alternative perspectives and inspire journalists to enrich their news content.
However, the results generated by Style-News can only serve as a reference and cannot be used directly.

\paragraph{Veracity of machine-generated news}
Following \cite{DBLP:conf/nips/ZellersHRBFRC19}, one of the goals of Style-News is to detect machine-generated news.
This task is necessary based on a strong assumption that machine-generated news is fake and can be harmful to the public.
Nonetheless, as we mention above, machine-generated news can also be regarded as a template or an inspiration for journalists. 
Therefore, we suggest that future work verify the factual claims of machine-generated news and release open-source datasets generated by different algorithms or researchers to construct stronger detectors.


\section{Limitations}
The major limitation of Style-News is the machine-generated news with further human modifications, i.e., multi-hop modifications.
The manual rewriting can be regarded as another various style, which increases the difficulty of neural fake news detection.
In addition, Style-News focuses on effective performance to mitigate the spread of neural fake news, but does not take the computation resource into account, which may be more efficient by introducing adapters into the model.

\section*{Acknowledgments}
We thank the anonymous reviewers for their insightful comments and feedback.
This work was partially supported by the Ministry of Science and Technology of Taiwan under Grants 112-2917-I-A49-007.

\bibliography{main}

\begin{thebibliography}{42}
\expandafter\ifx\csname natexlab\endcsname\relax\def\natexlab#1{#1}\fi

\bibitem[{Aich et~al.(2022)Aich, Bhattacharya, and
  Parde}]{DBLP:conf/coling/AichBP22}
Ankit Aich, Souvik Bhattacharya, and Natalie Parde. 2022.
\newblock Demystifying neural fake news via linguistic feature-based
  interpretation.
\newblock In \emph{{COLING}}, pages 6586--6599. International Committee on
  Computational Linguistics.

\bibitem[{Alkaissi and McFarlane(2023)}]{alkaissi2023artificial}
Hussam Alkaissi and Samy~I McFarlane. 2023.
\newblock Artificial hallucinations in chatgpt: implications in scientific
  writing.
\newblock \emph{Cureus}, 15(2).

\bibitem[{Baptista and Gradim(2020)}]{baptista2020understanding}
Jo{\~a}o~Pedro Baptista and Anabela Gradim. 2020.
\newblock Understanding fake news consumption: A review.
\newblock \emph{Social Sciences}, 9(10):185.

\bibitem[{Brown et~al.(2020)Brown, Mann, Ryder, Subbiah, Kaplan, Dhariwal,
  Neelakantan, Shyam, Sastry, Askell, Agarwal, Herbert{-}Voss, Krueger,
  Henighan, Child, Ramesh, Ziegler, Wu, Winter, Hesse, Chen, Sigler, Litwin,
  Gray, Chess, Clark, Berner, McCandlish, Radford, Sutskever, and
  Amodei}]{DBLP:conf/nips/2020}
Tom~B. Brown, Benjamin Mann, Nick Ryder, Melanie Subbiah, Jared Kaplan,
  Prafulla Dhariwal, Arvind Neelakantan, Pranav Shyam, Girish Sastry, Amanda
  Askell, Sandhini Agarwal, Ariel Herbert{-}Voss, Gretchen Krueger, Tom
  Henighan, Rewon Child, Aditya Ramesh, Daniel~M. Ziegler, Jeffrey Wu, Clemens
  Winter, Christopher Hesse, Mark Chen, Eric Sigler, Mateusz Litwin, Scott
  Gray, Benjamin Chess, Jack Clark, Christopher Berner, Sam McCandlish, Alec
  Radford, Ilya Sutskever, and Dario Amodei. 2020.
\newblock Language models are few-shot learners.
\newblock In \emph{NeurIPS}.

\bibitem[{Dathathri et~al.(2020)Dathathri, Madotto, Lan, Hung, Frank, Molino,
  Yosinski, and Liu}]{DBLP:conf/iclr/DathathriMLHFMY20}
Sumanth Dathathri, Andrea Madotto, Janice Lan, Jane Hung, Eric Frank, Piero
  Molino, Jason Yosinski, and Rosanne Liu. 2020.
\newblock Plug and play language models: {A} simple approach to controlled text
  generation.
\newblock In \emph{{ICLR}}. OpenReview.net.

\bibitem[{Devlin et~al.(2019)Devlin, Chang, Lee, and
  Toutanova}]{DBLP:conf/naacl/DevlinCLT19}
Jacob Devlin, Ming{-}Wei Chang, Kenton Lee, and Kristina Toutanova. 2019.
\newblock {BERT:} pre-training of deep bidirectional transformers for language
  understanding.
\newblock In \emph{{NAACL-HLT} {(1)}}, pages 4171--4186. Association for
  Computational Linguistics.

\bibitem[{Forbes(2023)}]{forbes2023}
Forbes. 2023.
\newblock Lawyer used chatgpt in court—and cited fake cases. a judge is
  considering sanctions.
\newblock
  https://www.forbes.com/sites/mollybohannon/2023/06/08/lawyer-used-chatgpt-in-court-and-cited-fake-cases-a-judge-is-considering-sanctions/?sh=3308159a7c7f.

\bibitem[{Fu et~al.(2020)Fu, Zhang, Meng, and King}]{DBLP:conf/www/0004ZMK20}
Xinyu Fu, Jiani Zhang, Ziqiao Meng, and Irwin King. 2020.
\newblock {MAGNN:} metapath aggregated graph neural network for heterogeneous
  graph embedding.
\newblock In \emph{{WWW}}, pages 2331--2341. {ACM} / {IW3C2}.

\bibitem[{Fung et~al.(2021)Fung, Thomas, Reddy, Polisetty, Ji, Chang, McKeown,
  Bansal, and Sil}]{DBLP:conf/acl/FungTRPJCMBS20}
Yi~Fung, Christopher Thomas, Revanth~Gangi Reddy, Sandeep Polisetty, Heng Ji,
  Shih{-}Fu Chang, Kathleen~R. McKeown, Mohit Bansal, and Avi Sil. 2021.
\newblock Infosurgeon: Cross-media fine-grained information consistency
  checking for fake news detection.
\newblock In \emph{{ACL/IJCNLP} {(1)}}, pages 1683--1698. Association for
  Computational Linguistics.

\bibitem[{Hermann et~al.(2015)Hermann, Kocisk{\'{y}}, Grefenstette, Espeholt,
  Kay, Suleyman, and Blunsom}]{DBLP:conf/nips/HermannKGEKSB15}
Karl~Moritz Hermann, Tom{\'{a}}s Kocisk{\'{y}}, Edward Grefenstette, Lasse
  Espeholt, Will Kay, Mustafa Suleyman, and Phil Blunsom. 2015.
\newblock Teaching machines to read and comprehend.
\newblock In \emph{{NIPS}}, pages 1693--1701.

\bibitem[{Higdon(2020)}]{breitbart}
Nolan Higdon. 2020.
\newblock The anatomy of fake news.
\newblock https://www.ucpress.edu/book/9780520347878/the-anatomy-of-fake-news.

\bibitem[{Huang et~al.(2022)Huang, Chen, and Chen}]{DBLP:conf/coling/HuangCC22}
Yen{-}Hao Huang, Yi{-}Hsin Chen, and Yi{-}Shin Chen. 2022.
\newblock Contexting: Granting document-wise contextual embeddings to graph
  neural networks for inductive text classification.
\newblock In \emph{{COLING}}, pages 1163--1168. International Committee on
  Computational Linguistics.

\bibitem[{Keskar et~al.(2019)Keskar, McCann, Varshney, Xiong, and
  Socher}]{DBLP:journals/corr/abs-1909-05858}
Nitish~Shirish Keskar, Bryan McCann, Lav~R. Varshney, Caiming Xiong, and
  Richard Socher. 2019.
\newblock {CTRL:} {A} conditional transformer language model for controllable
  generation.
\newblock \emph{CoRR}, abs/1909.05858.

\bibitem[{Kingma and Ba(2015)}]{DBLP:journals/corr/KingmaB14}
Diederik~P. Kingma and Jimmy Ba. 2015.
\newblock Adam: {A} method for stochastic optimization.
\newblock In \emph{{ICLR} (Poster)}.

\bibitem[{Lepp{\"{a}}nen et~al.(2017)Lepp{\"{a}}nen, Munezero,
  Granroth{-}Wilding, and Toivonen}]{DBLP:conf/inlg/LeppanenMGT17}
Leo Lepp{\"{a}}nen, Myriam Munezero, Mark Granroth{-}Wilding, and Hannu
  Toivonen. 2017.
\newblock Data-driven news generation for automated journalism.
\newblock In \emph{{INLG}}, pages 188--197. Association for Computational
  Linguistics.

\bibitem[{Li et~al.(2021)Li, Tang, Zhao, and
  Wen}]{DBLP:journals/corr/abs-2105-10311}
Junyi Li, Tianyi Tang, Wayne~Xin Zhao, and Ji{-}Rong Wen. 2021.
\newblock Pretrained language models for text generation: {A} survey.
\newblock \emph{CoRR}, abs/2105.10311.

\bibitem[{Liu et~al.(2021)Liu, Pillutla, Welleck, Oh, Choi, and
  Harchaoui}]{DBLP:conf/nips/LiuPWOCH21}
Lang Liu, Krishna Pillutla, Sean Welleck, Sewoong Oh, Yejin Choi, and
  Za{\"{\i}}d Harchaoui. 2021.
\newblock Divergence frontiers for generative models: Sample complexity,
  quantization effects, and frontier integrals.
\newblock In \emph{NeurIPS}, pages 12930--12942.

\bibitem[{Liu et~al.(2022)Liu, Zhang, Wang, Lan, and
  Shen}]{DBLP:journals/corr/abs-2212-10341}
Xiaoming Liu, Zhaohan Zhang, Yichen Wang, Yu~Lan, and Chao Shen. 2022.
\newblock Coco: Coherence-enhanced machine-generated text detection under data
  limitation with contrastive learning.
\newblock \emph{CoRR}, abs/2212.10341.

\bibitem[{Liu et~al.(2019)Liu, Ott, Goyal, Du, Joshi, Chen, Levy, Lewis,
  Zettlemoyer, and Stoyanov}]{DBLP:journals/corr/abs-1907-11692}
Yinhan Liu, Myle Ott, Naman Goyal, Jingfei Du, Mandar Joshi, Danqi Chen, Omer
  Levy, Mike Lewis, Luke Zettlemoyer, and Veselin Stoyanov. 2019.
\newblock Roberta: {A} robustly optimized {BERT} pretraining approach.
\newblock \emph{CoRR}, abs/1907.11692.

\bibitem[{OpenAI(2023)}]{DBLP:journals/corr/abs-2303-08774}
OpenAI. 2023.
\newblock {GPT-4} technical report.
\newblock \emph{CoRR}, abs/2303.08774.

\bibitem[{Papineni et~al.(2002)Papineni, Roukos, Ward, and
  Zhu}]{DBLP:conf/acl/PapineniRWZ02}
Kishore Papineni, Salim Roukos, Todd Ward, and Wei{-}Jing Zhu. 2002.
\newblock Bleu: a method for automatic evaluation of machine translation.
\newblock In \emph{{ACL}}, pages 311--318. {ACL}.

\bibitem[{Pegoraro et~al.(2023)Pegoraro, Kumari, Fereidooni, and
  Sadeghi}]{DBLP:journals/corr/abs-2304-01487}
Alessandro Pegoraro, Kavita Kumari, Hossein Fereidooni, and Ahmad{-}Reza
  Sadeghi. 2023.
\newblock To chatgpt, or not to chatgpt: That is the question!
\newblock \emph{CoRR}, abs/2304.01487.

\bibitem[{Pillutla et~al.(2021)Pillutla, Swayamdipta, Zellers, Thickstun,
  Welleck, Choi, and Harchaoui}]{DBLP:conf/nips/PillutlaSZTWCH21}
Krishna Pillutla, Swabha Swayamdipta, Rowan Zellers, John Thickstun, Sean
  Welleck, Yejin Choi, and Za{\"{\i}}d Harchaoui. 2021.
\newblock {MAUVE:} measuring the gap between neural text and human text using
  divergence frontiers.
\newblock In \emph{NeurIPS}, pages 4816--4828.

\bibitem[{Post(2018)}]{DBLP:conf/wmt/Post18}
Matt Post. 2018.
\newblock A call for clarity in reporting {BLEU} scores.
\newblock In \emph{{WMT}}, pages 186--191. Association for Computational
  Linguistics.

\bibitem[{Przybyla(2020)}]{DBLP:conf/aaai/Przybyla20}
Piotr Przybyla. 2020.
\newblock Capturing the style of fake news.
\newblock In \emph{{AAAI}}, pages 490--497. {AAAI} Press.

\bibitem[{Radford et~al.(2018)Radford, Narasimhan, Salimans, Sutskever
  et~al.}]{radford2018improving}
Alec Radford, Karthik Narasimhan, Tim Salimans, Ilya Sutskever, et~al. 2018.
\newblock Improving language understanding by generative pre-training.

\bibitem[{Radford et~al.(2019)Radford, Wu, Child, Luan, Amodei, Sutskever
  et~al.}]{radford2019language}
Alec Radford, Jeffrey Wu, Rewon Child, David Luan, Dario Amodei, Ilya
  Sutskever, et~al. 2019.
\newblock Language models are unsupervised multitask learners.
\newblock \emph{OpenAI blog}, 1(8):9.

\bibitem[{Reuters(2023)}]{reuters2023}
Reuters. 2023.
\newblock Google cautions against 'hallucinating' chatbots, report says.
\newblock
  https://www.reuters.com/technology/google-cautions-against-hallucinating-chatbots-report-2023-02-11/.

\bibitem[{See et~al.(2017)See, Liu, and Manning}]{DBLP:conf/acl/SeeLM17}
Abigail See, Peter~J. Liu, and Christopher~D. Manning. 2017.
\newblock Get to the point: Summarization with pointer-generator networks.
\newblock In \emph{{ACL} {(1)}}, pages 1073--1083. Association for
  Computational Linguistics.

\bibitem[{Shu et~al.(2021)Shu, Li, Ding, and Liu}]{DBLP:conf/aaai/ShuLDL21}
Kai Shu, Yichuan Li, Kaize Ding, and Huan Liu. 2021.
\newblock Fact-enhanced synthetic news generation.
\newblock In \emph{{AAAI}}, pages 13825--13833. {AAAI} Press.

\bibitem[{Shu et~al.(2017)Shu, Sliva, Wang, Tang, and
  Liu}]{DBLP:journals/sigkdd/ShuSWTL17}
Kai Shu, Amy Sliva, Suhang Wang, Jiliang Tang, and Huan Liu. 2017.
\newblock Fake news detection on social media: {A} data mining perspective.
\newblock \emph{{SIGKDD} Explor.}, 19(1):22--36.

\bibitem[{Syed et~al.(2020)Syed, Verma, Srinivasan, Natarajan, and
  Varma}]{DBLP:conf/aaai/SyedVSNV20}
Bakhtiyar Syed, Gaurav Verma, Balaji~Vasan Srinivasan, Anandhavelu Natarajan,
  and Vasudeva Varma. 2020.
\newblock Adapting language models for non-parallel author-stylized rewriting.
\newblock In \emph{{AAAI}}, pages 9008--9015. {AAAI} Press.

\bibitem[{Thompson(2018)}]{allthenews}
Andrew Thompson. 2018.
\newblock All the news dataset.
\newblock https://www.kaggle.com/datasets/snapcrack/all-the-news.

\bibitem[{Van~der Kaa and Krahmer(2014)}]{van2014journalist}
Hille Van~der Kaa and Emiel Krahmer. 2014.
\newblock Journalist versus news consumer: The perceived credibility of machine
  written news.
\newblock In \emph{Proceedings of the computation+ journalism conference,
  Columbia university, New York}, volume~24, page~25.

\bibitem[{Verma and Srinivasan(2019)}]{DBLP:journals/corr/abs-1909-08349}
Gaurav Verma and Balaji~Vasan Srinivasan. 2019.
\newblock A lexical, syntactic, and semantic perspective for understanding
  style in text.
\newblock \emph{CoRR}, abs/1909.08349.

\bibitem[{Wang et~al.(2019)Wang, Wu, Mou, Li, and
  Chao}]{DBLP:conf/emnlp/WangWMLC19}
Yunli Wang, Yu~Wu, Lili Mou, Zhoujun Li, and Wenhan Chao. 2019.
\newblock Harnessing pre-trained neural networks with rules for formality style
  transfer.
\newblock In \emph{{EMNLP/IJCNLP} {(1)}}, pages 3571--3576. Association for
  Computational Linguistics.

\bibitem[{Xu et~al.(2022)Xu, Wu, Liu, Wu, and Wang}]{DBLP:conf/www/XuWLWW22}
Weizhi Xu, Junfei Wu, Qiang Liu, Shu Wu, and Liang Wang. 2022.
\newblock Evidence-aware fake news detection with graph neural networks.
\newblock In \emph{{WWW}}, pages 2501--2510. {ACM}.

\bibitem[{Zellers et~al.(2019)Zellers, Holtzman, Rashkin, Bisk, Farhadi,
  Roesner, and Choi}]{DBLP:conf/nips/ZellersHRBFRC19}
Rowan Zellers, Ari Holtzman, Hannah Rashkin, Yonatan Bisk, Ali Farhadi,
  Franziska Roesner, and Yejin Choi. 2019.
\newblock Defending against neural fake news.
\newblock In \emph{NeurIPS}, pages 9051--9062.

\bibitem[{Zhang et~al.(2022)Zhang, Song, Li, Zhou, and
  Song}]{DBLP:journals/corr/abs-2201-05337}
Hanqing Zhang, Haolin Song, Shaoyu Li, Ming Zhou, and Dawei Song. 2022.
\newblock A survey of controllable text generation using transformer-based
  pre-trained language models.
\newblock \emph{CoRR}, abs/2201.05337.

\bibitem[{Zhang et~al.(2020)Zhang, Yu, Cui, Wu, Wen, and
  Wang}]{DBLP:conf/acl/ZhangYCWWW20}
Yufeng Zhang, Xueli Yu, Zeyu Cui, Shu Wu, Zhongzhen Wen, and Liang Wang. 2020.
\newblock Every document owns its structure: Inductive text classification via
  graph neural networks.
\newblock In \emph{{ACL}}, pages 334--339. Association for Computational
  Linguistics.

\bibitem[{Zhao et~al.(2019)Zhao, Peyrard, Liu, Gao, Meyer, and
  Eger}]{DBLP:conf/emnlp/ZhaoPLGME19}
Wei Zhao, Maxime Peyrard, Fei Liu, Yang Gao, Christian~M. Meyer, and Steffen
  Eger. 2019.
\newblock Moverscore: Text generation evaluating with contextualized embeddings
  and earth mover distance.
\newblock In \emph{{EMNLP/IJCNLP} {(1)}}, pages 563--578. Association for
  Computational Linguistics.

\bibitem[{Zhou and Zafarani(2021)}]{DBLP:journals/csur/ZhouZ20}
Xinyi Zhou and Reza Zafarani. 2021.
\newblock A survey of fake news: Fundamental theories, detection methods, and
  opportunities.
\newblock \emph{{ACM} Comput. Surv.}, 53(5):109:1--109:40.

\end{thebibliography}
\bibliographystyle{acl_natbib}

\appendix

\section{Experimental Setup}

\begin{table*}
  \small
  \centering
  \begin{tabular}{c|ccc}
  \toprule 
    Dataset & CNN/DailyMail & All the News & VOA-KG2txt \\
    \midrule \midrule
    Scenario & \makecell{\makecell{Synthetic News Generation, }\\ Publisher Prediction} & \makecell{\makecell{Synthetic News Generation, }\\ Publisher Prediction} & \makecell{Neural Fake News Detection} \\
    \midrule
    \# of train & 287,113 & 68,729 & 17,496 \\
    \# of validation & 13,368 & 2,988 & 6,006 \\
    \# of test & 11,490 & 2,989 & 5,782 \\
    \# of news & 311,971 & 74,706 & 29,284 \\
    \# of classes & 2 & 5 & 2 \\
    Avg. article length & 801.51 & 628.33 & 617.22 \\
    Avg. prompt length & 80.16 & 40.30 & 49.59 \\
    news highlight & summary & title & title \\
    \bottomrule
    \end{tabular}
\caption{The statistics of the datasets.}
\label{tab:stat}
\end{table*}

\subsection{Data Statistics}
The statistics of the datasets are described in Table \ref{tab:stat}.


\begin{figure}
    \centering
    \includegraphics[width=\linewidth]{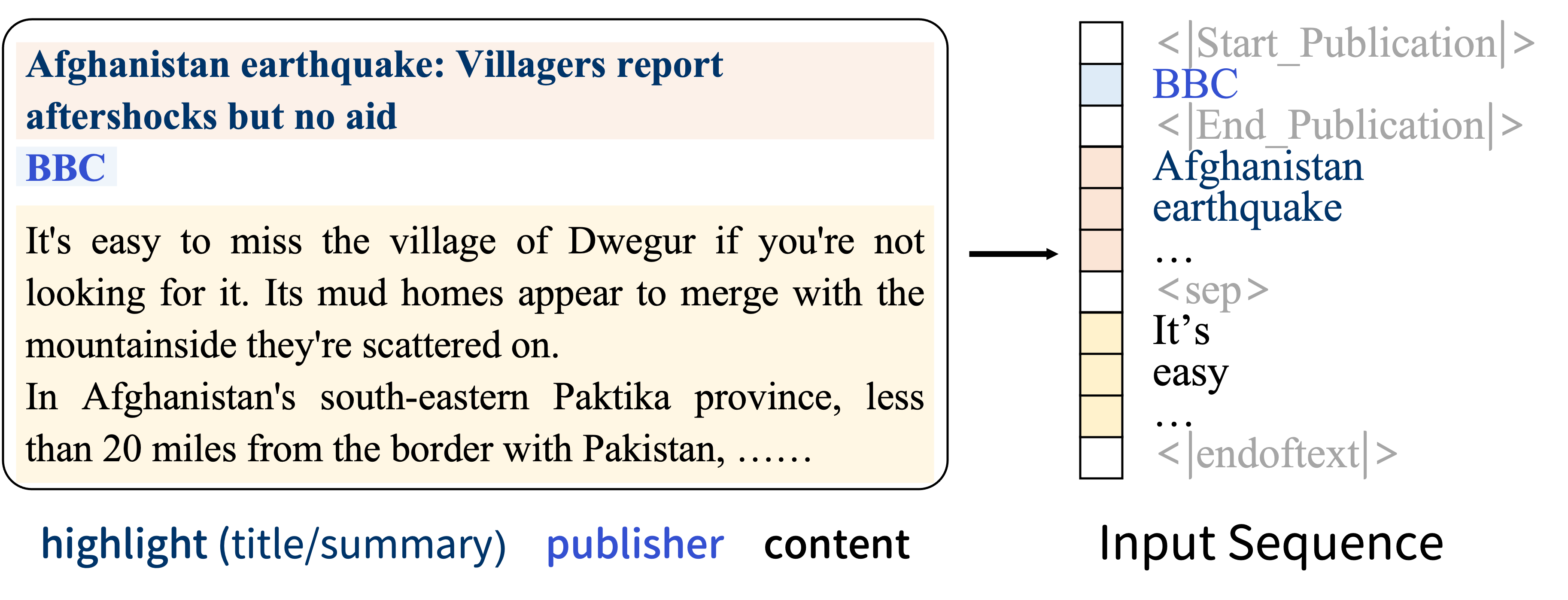}
    \caption{An illustration of the input prompt with news publisher, highlight, and content.}
    \label{Input_Sequence}
\end{figure}

\subsection{Details of Generative Experiments}
\subsubsection{Generative Baselines}
\label{appendix:generative-baselines}
The details of the generative baselines are depicted as follows:

We select 5 neural fake news baselines in this experiment to compare the generation quality of our Style-News.
All baselines were finetuned based on the corresponding dataset.
We note that only PPLM used publisher information during the experiment since it has an attribute classifier to take advantage of such information to jointly train the generator.
The others did not use publisher information to follow their original implementations.
\begin{itemize}
    \item[$\bullet$] CopyTransformer \cite{DBLP:conf/acl/SeeLM17}: a sequence-to-sequence model with a pointer generator and a coverage mechanism for abstractive summarization. 
    \item[$\bullet$] GPT-2 \cite{radford2019language}: a  foundation model on language tasks, which utilizes a combination of pre-training and supervised fine-tuning.
    \item[$\bullet$] PPLM$_{\textrm{gen}}$ \cite{DBLP:conf/iclr/DathathriMLHFMY20}: a controllable generative model, which combines a bag of words (BoW) and a discriminator attribute classifier to guide the language model.
    \item[$\bullet$] Grover$_{\textrm{gen}}$ \cite{DBLP:conf/nips/ZellersHRBFRC19}: a controllable generative model, which can produce neural fake news according to multi-field documents.
    \item[$\bullet$] FactGen \cite{DBLP:conf/aaai/ShuLDL21}: a synthetic news generative model, which focuses on factual discrepancies between the human-written and machine-generated text.
\end{itemize}

\subsubsection{Evaluation Metrics for Generated Results}
\label{appendix:generative-metrics}
The details of each metric are introduced as follows:
\begin{itemize}
    \item[$\bullet$] Language Fluency: To measure the quality of the machine-generated news, we introduce the Mauve score \cite{DBLP:conf/nips/PillutlaSZTWCH21} and Frontier Integral \cite{DBLP:conf/nips/LiuPWOCH21}, which compute quantized embeddings and produce divergence scores for evaluating the similarity of machine-generated text and human-written text.
    Both the Mauve score and Frontier Integral are between 0 and 1.
    A higher Mauve score and a lower Frontier Integral value indicate that the machine-generated text is closer to human-written text.
    We apply these two metrics to the news pairs \{$N_H, N_M$\} with the same metadata.
    \item[$\bullet$] Content Preservation: To validate the semantic similarity between the machine-generated text and the corresponding input prompt, we utilize SacreBLEU \cite{DBLP:conf/wmt/Post18} and MoverScore \cite{DBLP:conf/emnlp/ZhaoPLGME19}.
    SacreBLEU is an improved version of the BLEU \cite{DBLP:conf/acl/PapineniRWZ02} score by wrapping the BLEU score implementation together with useful features yielded from WMT\footnote{https://statmt.org/wmt17/}.
    MoverScore measures the distance of contextualized representations between machine-generated outputs and references.
    \item[$\bullet$] Style Adherence: To evaluate whether the generative model is able to capture the writing style of a given publisher, we trained RoBERTa-Large \cite{DBLP:journals/corr/abs-1907-11692} as the classification model on the basis of human-written news with the training sets of CNN/DailyMail and All the News, and achieved about 98\% and 90\% accuracy respectively on the validation set.
    During the evaluation, we adopted this trained publisher classifier to distinguish the publishers of machine-generated news content.
\end{itemize}

\subsection{Details of Discriminative Experiments}
\subsubsection{Discriminative Baselines}
\label{appendix:discriminative-baselines}

We select 5 discriminative baselines in this experiment to compare the discrimination quality of our Style-News:
\begin{itemize}
    \item[$\bullet$] RoBERTa \cite{DBLP:journals/corr/abs-1907-11692}: an optimized approach based on BERT \cite{DBLP:conf/naacl/DevlinCLT19}. 
    \item[$\bullet$] PPLM$_{\textrm{def}}$ \cite{DBLP:conf/iclr/DathathriMLHFMY20}: a linear discriminator trained on top of GPT-2 with pre-defined attributes to learn the style/sentiment classifications.
    \item[$\bullet$] Grover$_{\textrm{def}}$ \cite{DBLP:conf/nips/ZellersHRBFRC19}: a linear classifier trained to predict whether the input text is human-written or machine-generated. 
    \item[$\bullet$] GET \cite{DBLP:conf/www/XuWLWW22}: a unified graph-based semantic structure framework for information fusion via neighborhood propagation.
\end{itemize}

\label{shared_backbone}
\begin{table}
    \small
    \centering
    \begin{tabular}{cc|cc}
    \toprule
    Criteria & Metric & \textbf{Style-News} & \makecell{\textbf{Style-News}\\\textbf{Shared Backbone}} \\
    \midrule \midrule
    \multirow{2}{*}{Fluency} & Mauve & 0.8832 & \textbf{0.8845}\\
     & Frontier & 0.6734 & \textbf{0.6585}\\
    \midrule
    \multirow{2}{*}{Content} & SacreBLEU & \textbf{18.1064} & 15.537 \\
     & MoverScore & \textbf{0.5523} & 0.5523 \\
    \midrule
    \multirow{2}{*}{Style} & Accuracy & 0.9609 & \textbf{0.9630} \\
     & F1 & 0.8792 & \textbf{0.8877}\\
     \bottomrule
    \end{tabular}
    \caption{Comparison of synthetic news generation on CNN/DailyMail with or without a shared backbone.}
    \label{tab:app_cnndm}
\end{table}

\begin{table}
    \small
    \centering
    \begin{tabular}{cc|cc}
    \toprule
    Criteria & Metric & \textbf{Style-News} & \makecell{\textbf{Style-News}\\\textbf{Shared Backbone}} \\
    \midrule \midrule
    \multirow{2}{*}{Fluency} & Mauve & 0.8881 & \textbf{0.8932} \\
    & Frontier & 0.6467 & \textbf{0.6259} \\
    \midrule
    \multirow{2}{*}{Content} & SacreBLEU & \textbf{14.5186} & 11.0700 \\
    & MoverScore & \textbf{0.5448} & 0.5371 \\
    \midrule
    \multirow{2}{*}{Style} & Accuracy & \textbf{0.5937} & 0.5681 \\
    & F1 & \textbf{0.4921} & 0.4496 \\
    \bottomrule
    \end{tabular}
    \caption{Comparison of synthetic news generation on All the News with or without a shared backbone.}
    \label{tab:app_atn}
\end{table}

\subsubsection{Evaluation Metrics for Discriminative Results}
\label{appendix:discriminative-metrics}
\begin{itemize}
\item[$\bullet$] Accuracy: the proportion of the number of correct predictions over total samples.
\item[$\bullet$] Precision: the proportion of the number of correct positive predictive values over all positive samples.
\item[$\bullet$] Recall: the proportion of the number of correct positive predictive values over all correct samples.
\item[$\bullet$] F1 Score: the harmonic mean of precision and recall, which help keep a balance among these two metrics. 
\end{itemize}

\section{Additional Experiments}
\subsection{Shared Discriminators of Style and Source}
Since the discriminative mechanism (DM) design is the same as for the style discriminator and source discriminator, we conducted an additional experiment on the shared backbone of these two discriminators to observe the effect of sharing the weights.
As shown in Tables \ref{tab:app_cnndm} and \ref{tab:app_atn}, we can observe that the shared backbone performs slightly better in the CNN/DailyMail in terms of language fluency and style adherence, while the performance drops obviously on content preservation and style adherence in All the News, which indicates that the lightning parameters of discriminators (shared backbone) might improve in certain cases.

\begin{figure}[]
    \centering
    \begin{subfigure}[b]{0.23\textwidth}
         \centering
         \includegraphics[width=\linewidth]{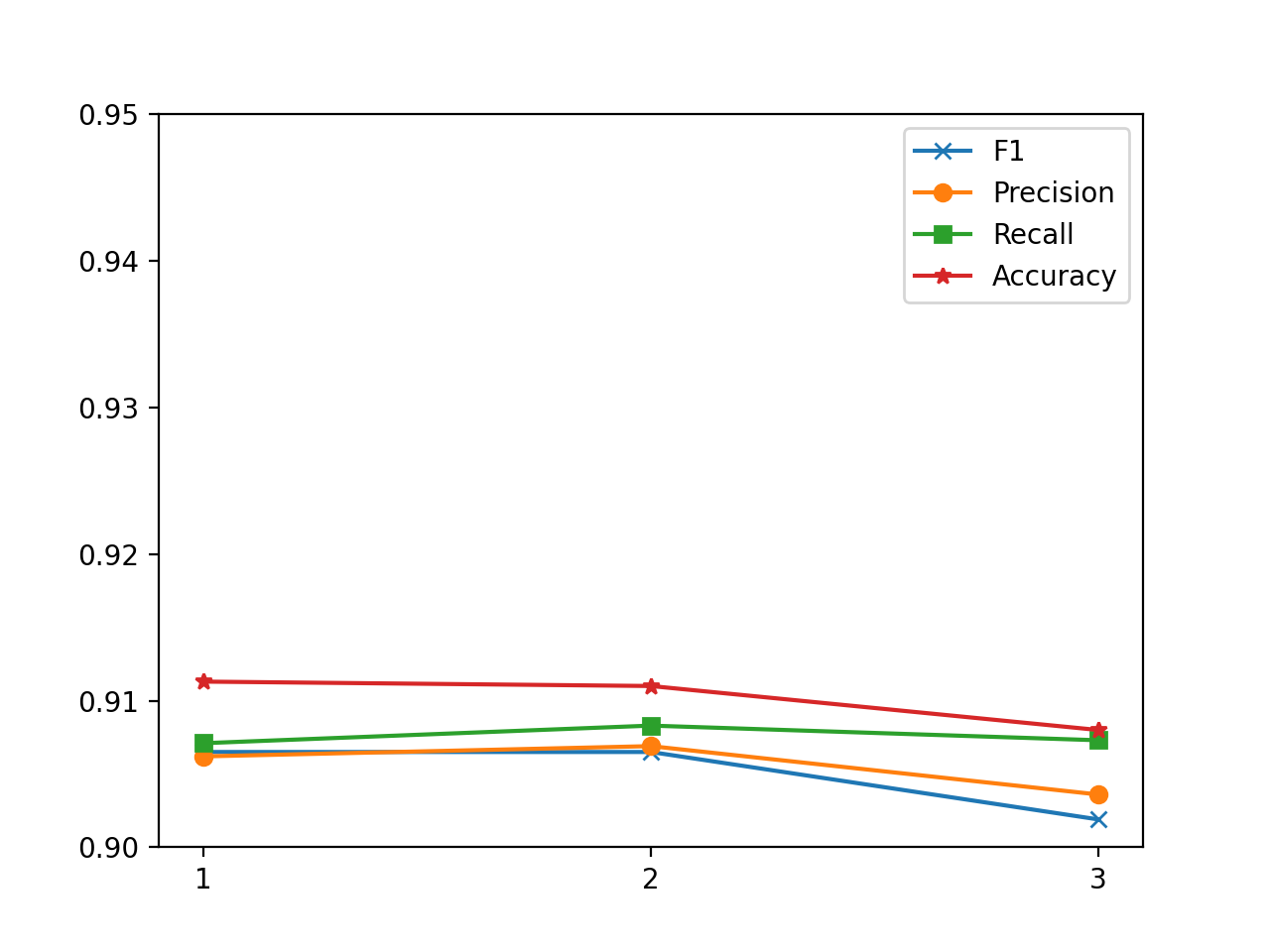}
         \caption{CNN/DailyMail}
    \end{subfigure}
    \begin{subfigure}[b]{0.23\textwidth}
         \centering
         \includegraphics[width=\linewidth]{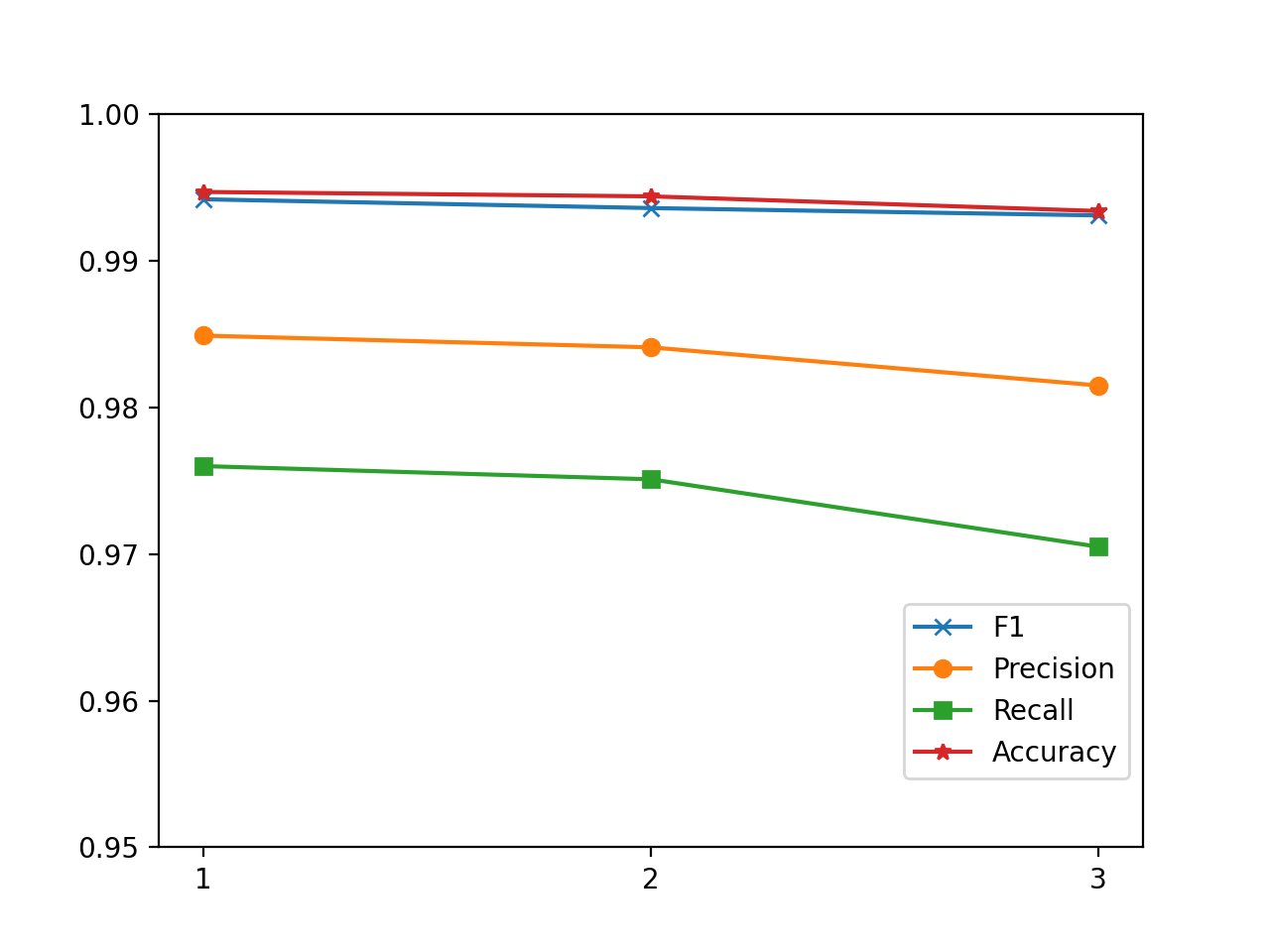}
         \caption{All the News}
    \end{subfigure}
    \caption{Analysis of the number of hops.}
    \label{fig:hop-sensitivity}
\end{figure}

\subsection{Parameter Sensitivity}
To investigate the effects of the number of hops $p$ in DM, Figure \ref{fig:hop-sensitivity} demonstrates the performance of each metric in terms of CNN/DailyMail and All the News datasets with $p$ being \{1, 2, 3\}.
We can observe that the performance is stable between different numbers of hops in terms of F1, precision, recall, and accuracy metrics, where all settings still outperform other baselines, which signifies the robust design of the inductive word graph in DM.




\subsection{Case Study}
\label{case}

\subsubsection{Generated Examples with Different Methods}
We further illustrate the generated samples by different models as shown in Table \ref{tab:gen_case}.
The observations are summarized as follows:
1) We can see that the results of CopyTransformer and FactGen tend not to be like a human since there are <unk> tokens and repetitive words generated in an unusual way.
2) Although Grover can generate fluent sentences, the discussion topic is totally different from the given topic. 
3) Style-News not only preserves the original meaning of the input prompt, but also explores an issue on \redit{human rights}, which also appears in human-written content.
4) This news actually discusses the standpoint of the United States; however, in Style-News, the subject is replaced with the British government. As the misinformation is intentionally or accidentally generated, it can be spread and mislead the public without confirmation, which causes the risk of neural fake news detection.
Therefore, our source discriminator is able to detect it as neural fake news.
This case study shows the generation results from different models, which expose the possible success of exploration on relevant issues and the potential risk of altering political stance that leads to information warfare.

\begin{table}[b]
    \small
    \begin{tabular}{c|cc|cc}
\toprule
Publisher & \multicolumn{2}{c|}{CNN} & \multicolumn{2}{c}{DailyMail} \\
Spelling  & British & American & British & American \\
\midrule
Occurrence & 161 & 690 & 932 & 179 \\
\bottomrule
\end{tabular}
    \caption{The occurrences of using different publishers as prompts.}
    \label{tab:case-occ}
\end{table}

\begin{table*}[t]
    \centering
    \begin{tabular}{Q{4cm}|Q{11cm}} 
    \toprule
    Title & Human-written content \\
    \midrule \midrule
    \uline{Membership gives the ICC jurisdiction over alleged crimes committed in Palestinian territories since last June.}  & \uline{Israel and the United States opposed the move, which could open the door to war crimes investigations against Israelis.} ... Rights group \redit{Human Rights} Watch welcomed the development . " Governments seeking to penalize Palestine for joining the \textbf{ICC} should immediately end their pressure , and countries that support universal acceptance of the court 's treaty should speak out to welcome its membership , " said Balkees Jarrah , international justice counsel for the group. ... \\
    \bottomrule
    \end{tabular}
    \begin{tabular}{Q{2.8cm}|Q{12.2cm}} 
    \toprule
    Model & Generation result \\
    \midrule \midrule
    CopyTransformer & <unk> where . in \textbf{jurisdiction} for , accession , Netherlands , <unk> CNN <unk> was not guilty in the the 123rd , but had been step a accession \'s a child he will be a when the founding has do more to Hague \textbf{Palestinian} from that the top wife with \textbf{crimes} Netherlands  formal was an the formal \textbf{territory} .\\
    \midrule
    GPT-2 & ... \textbf{ICC}'s decision rests mainly at home, but may become more severe, especially when an \textbf{ICC} opinion outside the court system leads from overseas where disputes between partners who challenge legitimate authority should prevail rather than within their own borders remain undemocratic. ...\\
    \midrule
    PPLM$_{\textrm{gen}}$ & \textbf{Palestinian} President Mahmoud Abbas is expected to announce the new status of a \textbf{territory} he has occupied since 1967 . " The international community has to stop supporting this occupation ," Abbas said at an official news conference in Ramallah . The \textbf{ICC} could investigate cases of war \textbf{crimes} committed during \textbf{Israel}'s war against Hamas and \textbf{Palestinian} fighters in the Gaza Strip in 2006 ... \\
    \midrule
    Grover$_{\textrm{gen}}$ & A transgender mother of three is speaking out for the first time about her torment ordeal of sex and gender identity issues after a New York attorney gave him a ride on the bus . The woman who suffers from gender and gender issues is shocked and has a boyfriend , say the lawyer , David Kesser . ...\\
    \midrule
    FactGen & ... Members could also \textbf{opposed} war over in \textbf{Palestinian territories} and \textbf{Palestinian territories} in \textbf{territories}  overhip could move to war to \textbf{Israelis} . and . \textbf{Israel} gives the \textbf{ICC jurisdiction} over '\textbf{Israel} and \textbf{jurisdiction} over \textbf{crimes} , opposed in alleged \textbf{Palestinian territories} since \textbf{June} \textbf{June} . . over and \textbf{United States} . \textbf{last June} . in \textbf{June last} ... \\
    \midrule
    \textbf{Style-News} & The British government is concerned about allegations of abuse by Gaza militants, a possible rise on accusations from \textbf{Palestinian} prisoners of terrorism to drug - dealing \textbf{crimes} involving \textbf{Israelis} who have come a close behind. ... There's also concern that \redit{human rights} organizations may need additional information, ... \\
    \bottomrule
    \end{tabular}
\caption{Case study of the machine-generated result on CNN/DailyMail. The input prompt for models is the token sequence containing the publisher, title, and the first sentence of human-written content, which is \underline{underlined}. The sentence fragment marked in \textbf{black bold} is consistent information with the input prompt, and the \redit{red italics} is the supplementary information.}
\label{tab:gen_case}
\end{table*}

\subsubsection{Different Styles of Generated News}

\begin{table*}
    \begin{tabular}{Q{2.8cm}|Q{12.2cm}} 
\toprule
Publisher & Generated result \\
\midrule
DailyMail & Angus was - the name for James Hawke. The father of three died two years ago. ... Evidence compiled for coroner Anthony Harkin shows several major organ donor donors had undergone surgery to reduce complications in each set where their organs appeared healthy enough prior diagnosis as part treatment with standard IV catheterisation following blood transfusions, known collectively under \textbf{UK protocol} here \textbf{NHS.gov} works closely behind EU protocols designed specifically place high levels above donors health records ( file picture ) One individual identified only locally as Michael'may wish themselves happy knowing this process has been completed without any involvement whatsoever, " according Australian law firm Deverer \& Wilshere ( WRW ). They claim further evidence has indicated poor screening among those donating organs in \textbf{Europe}. ... \\ 
\midrule
CNN       & Angus was a - tonne fighter pilot and captain on the USS Cheyenne before joining Operation Blue Angels, he told \textbf{CNN Sunday} night about plans for one of the ships that were sunk by an F / A-18 Hornet earlier this month at Sea Wolf Reef off Alaska province ; he died six weeks later when a ship sank. ... After being accused of involvement then serving 13 months behind bars in prison near Sydney and at Guantanamo Bay in 2001 awaiting trial on charges unrelated terrorism offenses for shooting dead three \textbf{US embassy} workers September 11 in Pakistan. ...  \\
\bottomrule
\end{tabular}
    \caption{The sampled result of using different publishers.}
    \label{tab:case-style-difference}
\end{table*}

Since the concept of style is relatively abstract, we follow \cite{DBLP:conf/aaai/Przybyla20} to use statistical methods to quantify the style differences on the CNN/DailyMail dataset.
Specifically, CNN and DailyMail are famous media in the US and UK respectively; therefore, we use American and British spellings (e.g., \textit{analyze} and \textit{analyse}) to analyze different styles corresponding to given authors.
Table \ref{tab:case-occ} reports the total occurrence of presenting American and British spellings on the test set.
We can observe that the occurrence of British spelling is 5.78 more often than that of American spelling if the publisher is set to DailyMail.
On the other hand, the term frequency becomes 0.26 times if the given publisher is changed to CNN.

We also use a sampled case of generated results to study the qualitative difference between the style of CNN and DailyMail.
We hypothesize that the implications of national culture and social background can be one of the aspects for explicitly qualifying different styles.
Given the same description \textit{Angus Hawley's brother said his late sibling' didn't have heart problems. He is reported to have had a suspected heart attack in New York.} as a prompt and modifying the publisher to be either CNN or DailyMail, as shown in Table \ref{tab:case-style-difference}, we observe that the generated result of the DailyMail mentions \textbf{UK protocol}, \textbf{NHS.gov}, and \textbf{Europe}.
Besides, CNN's generated result mentions \textbf{CNN Sunday}, and the \textbf{US embassy}.
These studies show that publisher information causes Style-News to generate different content regarding the given news, and thus publishers can potentially be utilized to control neural models to produce malicious news.

\end{document}